\documentclass[letterpaper, 10 pt, conference]{ieeeconf}  

\IEEEoverridecommandlockouts                              

\usepackage{amsmath}
\usepackage[linesnumbered,ruled]{algorithm2e}
\usepackage{graphicx}
\usepackage{balance}
\begin{document}

\title{ \LARGE \bf End-to-End Semi-supervised Learning for Differentiable
Particle Filters}

\author{Hao Wen,
        Xiongjie Chen,
        Georgios Papagiannis,
        Conghui Hu,
        and~Yunpeng Li
\thanks{The authors are with Department of Computer Science, University of Surrey, Surrey GU2 7XH, United Kingdom, e-mail:\{h.wen, xiongjie.chen, g.papagiannis, conghui.hu, yunpeng.li\}@surrey.ac.uk.}
}

\maketitle

\begin{abstract}\label{sec:abstract}
Recent advances in incorporating neural networks into particle filters provide the desired flexibility to apply particle filters in large-scale real-world applications. The dynamic and measurement models in this framework are learnable through the differentiable implementation of particle filters. Past efforts in optimising such models often require the knowledge of true states which can be expensive to obtain or even unavailable in practice. In this paper, in order to reduce the demand for annotated data, we present an end-to-end learning objective based upon the maximisation of a pseudo-likelihood function which can improve the estimation of states when large portion of true states are unknown. We assess performance of the proposed method in state estimation tasks in robotics with simulated and real-world datasets.
\end{abstract}

\section{Introduction}\label{sec:introduction}
    
    Sequential state estimation task, which involves estimating unknown state from a sequence of observations, finds a variety of applications including target tracking \cite{zhang2017multi, dai2019visual}, navigation \cite{wang2017unbiased, palmier2019adaptive}, and signal processing \cite{qian20173d, liu2019audio}. Recursive Bayesian filtering is a probabilistic approach to sequential state estimation \cite{doucet2001introduction}, which requires the specification of a dynamic model that describes how the hidden state evolves over time and a measurement model that defines the likelihood of the observation given the estimated state.
    
    Algorithms for solving such a problem involve constructing these probabilistic models or learning them from data
    \cite{hurzeler2001approximating,liu2001combined}. Estimation of the models' parameters is often known as parameter inference,
    which normally comprises either Bayesian or Maximum Likelihood (ML) methods~\cite{kantas2009overview, kantas2015particle}.
    These methods can also be classified as offline or online depending on whether the data is processed in batch \cite{poyiadjis2005maximum,malik2011particle} or recursively \cite{kantas2015particle,andrieu2005line}.
    Online approach is often preferable with streaming data, especially if the amount of data to be processed is large~\cite{andrieu2005line}.
    While online Bayesian methods suffer from degeneracy problems, maximum likelihood-based methods, including direct optimisation approaches, gradient ascent, Expectation Maximisation (EM), have shown promises in parameter inference with large datasets~\cite{kantas2015particle}. 
    
    However, the aforementioned methods assume that the structures or part of parameters of the dynamic and measurement models are known \cite{kantas2015particle}. Yet, formulating the dynamic model and measurement model is challenging, especially when the hidden state and observation data are in high-dimensional space or there is a lack of prior knowledge about the dynamic and measurement models. For example, sophisticated models are often required to describe the characteristics exhibit in complex, heterogeneous datasets such as streaming videos, text and medical records \cite{naesseth2018variational}. For these tasks, it is often impractical to specify dynamic and measurement models a priori \cite{coskun2017long}. 
    
    Recent advances in incorporating neural networks into Bayesian filtering provide the desired flexibility to learn the dynamic and measurement models and their parameters. Traditional Bayesian filtering methods can be implemented as differentiable filters, i.e. the dynamic and measurement models can be specified through neural networks and their parameters can be learned through backpropagation. Differentiable filters have been implemented for Kalman filters \cite{krishnan2015deep,haarnoja2016backprop,coskun2017long}, histogram filters \cite{jonschkowski2016end}, and particle filters~\cite{jonschkowski2018differentiable, karkus2018particle, ma2020particle, kloss_rss_ws_2020}. The incorporation of neural networks to particle filters, a.k.a. sequential Monte Carlo methods, is especially appealing due to the nonlinear nature of
    neural networks and the ability of particle filters to track posterior distribution with non-linear non-Gaussian models. However, to the best of our knowledge, optimisation of all of these differentiable filters requires sufficient training data with true state information or labels during the training stage, which can be expensive to obtain or even inaccessible in real-world scenarios. 
    
    To this end, we present semi-supervised differentiable particle filters (SDPFs) which leverage the unlabelled data in addition to a small subset of labelled data samples. This semi-supervised learning setting is of immense practical interest for differentiable particle filters in a wide range of applications where the unlabelled data are abundant and the access to labels is limited \cite{yousif2015overview,zhai2019s4l,berthelot2020remixmatch}. Our main contributions are three-folds: (i) We introduce an end-to-end learning objective based upon the maximisation of a pseudo-likelihood function for semi-supervised parameter inference in the differentiable particle filtering framework; (ii) We specify general dynamic and measurement model structures that can facilitate the optimisation of differentiable particle filters while maintaining the flexibility of models; (iii) We demonstrate superior performance of the proposed  semi-supervised differentiable particle filters through experiments with simulated and real-world datasets.

    The rest of this paper is organised as follows. We present the problem statement in Section~\ref{sec:ps}. A discussion of the background and related work is provided in Section~\ref{sec:background}. We introduce the proposed method in
    Section~\ref{sec:method}. Section~\ref{sec:experiment} provides details of experiments to evaluate the performance of proposed method. We conclude the paper in Section~\ref{sec:conclusion}.

\section{Problem statement}
\label{sec:ps}

    Recursive Bayesian filtering involves tracking the marginal posterior distribution $p(s_t|o_{1:t},a_{1:t})$, where $s_t$ is the hidden state at current time $t$, $o_{1:t} = \{o_1, o_2, \ldots, o_t\}$ and $a_{1:t} = \{a_1, a_2, \ldots, a_t\}$ are the history of observations and actions up to time step $t$, respectively.
    Bayesian filtering is considered in general probabilistic state space models where the dynamic and measurement models can be described as:
    \begin{align}
	    s_1&\sim \mu(s_1)\,\,, \\
	    s_t&\sim g_\theta(s_t| s_{t-1},a_t) \text{ for } t\geq2\,\,, \label{eq:transition} \\
        o_t&\sim l_\theta(o_t| s_t) \text{ for } t\geq2 \,\,\label{eq:likelihood}.
    \end{align}
	$\mu(s_1)$ is the stationary distribution of the hidden state at initial time, $g_\theta(s_t| s_{t-1},a_t)$ is the dynamic model which describes the transition of hidden state. $l_\theta(o_t| s_t)$ is the measurement model which describes the relation between the observation $o_t$ and the hidden state $s_t$. We use $\theta$ to denote all parameters of the dynamic and measurement models. In this paper, we focus on estimating the parameters $\theta$ of the dynamic and measurement models parameterised by neural networks.

\section{Background and related work}
\label{sec:background}

In this section, we briefly review how the bootstrap particle filter and differentiable particle filters are used for recursive Bayesian filtering.
\subsection{Bootstrap particle filter}

	Particle filters estimate the posterior distribution of hidden state by a set of weighted samples $\{(s_t^i,w_t^i)\}_{i=1}^{N_p}$, where $s_t^i$ is the state value of the $i$-th particle at time step $t$, $w_t^i$ is the corresponding particle weight and $\sum_{i=1}^{N_p} w_t^i=1$,
	and $N_p$ represents for the number of particles. In the bootstrap particle filter \cite{gordon1993novel}, the dynamic model and measurement model are used to propagate the particles and update the associated weights, respectively, which can be described by the following models:
	\begin{align}
	    s_1^i&\sim \mu(s_1)\,\,, \label{eq:stationary}\\
        \tilde{s}_t^i&\sim g_\theta(\tilde{s}_t| s_{t-1}^i,a_t) \text{ for } t\geq2 \,\,,\label{eq:dynamics}\\
        l_t^i&= l_\theta(o_t| \tilde{s}_t^i) \text{ for } t\geq2 \,\,,\label{eq:measurement}
    \end{align}
	where $g_\theta(\cdot|\cdot,\cdot)$ is the dynamic model which describes the transition of the hidden state. $\tilde{s}_t^i$ is the predicted particle state for the $i$-th particle at time step $t$ and $l_t^i$ is the observation likelihood. $\theta$ denotes all parameters of the dynamic and measurement models. The particle weights are updated recursively according to the observation likelihood:
	\begin{equation}
	\label{eq:weight_update}
	    w_t^i=\zeta  w_{t-1}^i l_t^i\,\,,
	\end{equation}
	where $\zeta$ is the normalization factor and $\zeta^{-1}=\sum_{i=1}^{N_p} w_{t-1}^i l_t^i$. Particles can be resampled proportional to their weights to avoid particle degeneracy.

\subsection{Differentiable particle filters}
	
	In most variants of particle filters, the dynamic model $g_\theta(\tilde{s}_t| s_{t-1}^i,a_t)$ and measurement model $l_\theta(o_t| \tilde{s}_t^i)$ are assumed to be known. In order to remove this restriction,  differentiable particle filters \cite{jonschkowski2018differentiable,karkus2018particle,ma2020particle} are proposed to learn the parameters $\theta$ of models $g_\theta(\tilde{s}_t| s_{t-1}^i,a_t)$ and $l_\theta(o_t| \tilde{s}_t^i)$ from data. Differentiable particle filters incorporate neural network into Bayesian inference where the learnable dynamic and measurement models can be optimised through gradient descent. As the sampling and resampling operations are not differentiable, the \emph{reparameterization trick} was proposed
	in \cite{kingma2013auto} to perform the sampling operation in a differentiable fashion. Specifically, the reparameterization trick constructs the prior distribution of the hidden state by employing a deterministic and differentiable function w.r.t its input and a noise vector. 
	
    The architecture design of dynamic and measurement models for differentiable particle filters varies in different applications. 
    For the design of the dynamic model, in  \cite{jonschkowski2018differentiable} and \cite{karkus2018particle}, the relative motion of hidden state is learned by transforming actions in the local coordinate to the global coordinate through a neural network. For measurement model, \cite{jonschkowski2018differentiable} adopts a neural network with observation features as input which directly outputs the likelihood of predicted particles. In \cite{karkus2018particle}, the measurement model is learned by incorporating the features of observation and a global map. In \cite{ma2020particle}, both dynamic and measurement models are learned using the recursive neural network (RNN) architecture. 
    
	In order to achieve differentiable resampling, the backpropagation of gradient is truncated during resampling at every time step \cite{jonschkowski2018differentiable}. Besides, \cite{karkus2018particle} introduced \emph{soft resampling} such that the gradient can be backpropagated from the weight of the sampled ancestors. In \cite{zhu2020towards}, a learned neural network resampler, which transforms the particle sets by a permutation-invariant encoder-decoder network, was introduced and resampled particles are aggregated by pooling.
	
	The optimisation of existing differentiable particle filters \cite{jonschkowski2018differentiable,karkus2018particle,ma2020particle} relies on the knowledge of true states for all time steps. In \cite{jonschkowski2018differentiable}, the parameters of the differentiable particle filters are learned by minimising the negative log-likelihood (NLL) of the true state under the estimated posterior distribution. In~\cite{karkus2018particle} and \cite{ma2020particle}, differentiable particle filters are trained by minimising the mean squared error (MSE) between the predicted state and the ground truth state.

\section{Semi-supervised learning for differentiable particle filters}
\label{sec:method}

   To alleviate the requirements for annotated data, we formulate a semi-supervised learning framework for training differentiable particle filters and introduce specifically designed dynamic and measurement models to facilitate the learning procedure.
    
    \textbf{Dynamic model.}
For efficient and stable gradients calculation between time steps, we adopt the practice in \cite{jonschkowski2018differentiable} and \cite{karkus2018particle} where
the relative motion of hidden state between time steps is modelled. The dynamic model is formulated as:
    \begin{equation}
	\label{eq:dynamics_semi}
	    \tilde{s}_t^i\sim g_\theta(\tilde{s}_t| s_{t-1}^i,a_t)=s_{t-1}^i+f_\theta(s_{t-1}^i, a_t)+\epsilon^i\,\,,
	\end{equation}
	where $\theta$ denotes the parameters of a neural network $f_\theta$ which transforms actions in the local coordinate to the global coordinate and models the relative motion between time steps, and $\epsilon^i$ is an auxiliary noise vector with independent marginal $p(\epsilon)$ used in the reparameterization trick \cite{kingma2013auto}.
    
    \textbf{Measurement model.}
    The measurement model in Eq. \eqref{eq:measurement} defines the likelihood of the observation given the predicted state. The measurement model can be regarded as a generative model which provides the distribution of observation data. However, if the observation data are in high-dimensional spaces, it is challenging to directly formulate the measurement model effectively. In such cases, an encoder network can be used to learn a more compact representation of observation data for dimension reduction. We approximate the likelihood of the observation by calculating the similarity between the observation data $o_t$ and the predicted state $\tilde{s}_t$ in a learned feature space. Specifically, an observation encoder $h_\theta: o_t \mapsto e_t$ implemented with a neural network is used to compress the observation $o_t$ into a feature vector $e_t$, and another neural network $\hat{h}_\theta: \tilde{s}_t \mapsto \tilde{e}_t$, which extracts features $\tilde{e}_t$ from predicted particle states $\tilde{s}_t$, is employed as the state encoder. In summary, the measurement model $l_\theta(o_t| \tilde{s}_t^i)$ can be expressed as follows:
    \begin{align}
        e_t &= h_\theta (o_t)\,\,, \\
        \tilde{e}_t^i &= \hat{h}_\theta ( \tilde{s}_t^i)\,\,,\\
        l_t^i&= l_\theta(o_t| \tilde{s}_t^i) = d_\theta (\tilde{e}_t^i, e_t)\,\,, \label{eq:measurement_semi}
    \end{align}
    where $d_\theta (\tilde{e}_t^i, e_t)$ measures the similarity between the features $\tilde{e}_t^i$ and $e_t$, and higher likelihood $l_t^i$ implies smaller distance between $\tilde{e}_t^i$ and $e_t$. Specifically, $d_\theta (\tilde{e}_t^i, e_t)$ is defined as:
    \begin{align}
        d_\theta (\tilde{e}_t^i, e_t) &= 1 / c (\tilde{e}_t^i, e_t)\,\,,\\
        c (\tilde{e}_t^i, e_t) &= 1 - \frac{\tilde{e}_t^i \cdot e_t}{||\tilde{e}_t^i||_2 || e_t||_2 }\,\,,
    \end{align}
    where $c (\tilde{e}_t^i, e_t)$ calculates the cosine distance between $\tilde{e}_t^i$ and $e_t$.
    
    \textbf{Semi-supervised learning.} We derive a learning objective for semi-supervised learning based on maximising a pseudo-likelihood. We start by defining a logarithmic pseudo-likelihood $Q_{semi}(\theta)$:
    \begin{equation}
    Q_{semi}(\theta)=\sum_{b=0}^{m-1}\log p_\theta(O_b|A_b)\,\,,
    \end{equation}
    where $O_b=o_{bL+1:(b+1)L}$ and $A_b=a_{bL+1:(b+1)L}$ are the $b$-th block of observations and the $b$-th block of actions respectively. $m$ denotes the number of blocks and $L$ is the block length.  
    The likelihood of the $b$-th block of observations $p_\theta(O_b|A_b)$ can be expressed by:
    \begin{equation}
        \begin{split}
            p_\theta(O_b|A_b)&=\int_{s^L}p_\theta(S_b,O_b|A_b)dS_b\\
            &=\int_{s^L}p_\theta(S_{b}|A_{b})p_\theta(O_{b}|S_{b},A_{b})dS_b\,\,,
        \end{split}
    \end{equation}
    where $S_b=s_{bL+1:(b+1)L}$ is the $b$-th block of states.
    
    Due to the Markovian property of the dynamic model as in Eq. \eqref{eq:transition} and the measurement model described via Eq. \eqref{eq:likelihood}, $p_\theta(S_{b}|A_{b})$ and $p_\theta(O_{b}|S_{b},A_{b})$ can be expressed as:
    \begin{equation}
        p_\theta(S_{b}|A_{b})=\mu_\theta({s}_{bL+1})\prod_{m=bL+2}^{(b+1)L}g_\theta({s}_m| s_{m-1},a_m)\,\,,
    \end{equation}
    \begin{equation}\label{eq:block_likelihood}
        p_\theta(O_{b}|S_{b},A_{b})=\prod_{m=bL+1}^{(b+1)L}l_\theta(o_{m}|{s}_{m})\,\,.
    \end{equation}
    Eq. \eqref{eq:block_likelihood} is due to Eq. \eqref{eq:likelihood} where $o_m$ and $a_m$ are conditionally independent given $s_m$ for a given time step $m$.

    According to the ergodicity assumption in employing pseudo-likelihood \cite{andrieu2005line}, the average of logarithmic pseudo-likelihood satisfies:
    \begin{equation}
        \lim_{m\to\infty}\frac{1}{m}\sum_{b=0}^{m-1}\log p_\theta(O_b|A_b)=\bar{Q}_{semi}(\theta)\,\,,
    \end{equation}
    where $\bar{Q}_{semi}(\theta)$ is defined as:
    \begin{equation}
        \bar{Q}_{semi}(\theta)=\int_{o^L}\log p_\theta(O|A)\cdot p_{\theta^*}(O|A)d O\,\,.
    \end{equation}
    $\theta^*$ denotes the optimal parameter value. $O$ and $A$ are a block of observations and actions. The parameter $\theta$ can be updated recursively by:
    \begin{equation}
        \theta_{b+1}=\underset{\theta\in\Theta}{\arg\max}\,Q(\theta,\theta_{b})\,\,,
    \end{equation}
    where
    \begin{equation}
    \begin{split}
        Q(\theta,\theta_{b})=&\int_{s^L\times o^L}\log(p_\theta(S,O|A))p_{\theta_{b}}(S|O,A)\\&p_{\theta^*}(O|A)dSdO,
    \end{split}
    \end{equation}
    and $S$ is a block of states, and $\theta_b$ is the parameter value estimated at the $b$-th block. 
    
    With the differentiable particle filtering framework, we can approximate $Q(\theta,\theta_{b})$ with $\hat{Q}(\theta,\theta_{b})$ as follows to form
    the optimisation objective for samples without true labels:
    \begin{equation}\label{eq:Q}
        \begin{split}
            \hat{Q}(\theta,\theta_{b})&=\sum_{i=1}^{N_p}w_{b}^i\log p_\theta(S_{b}^i,O_{b}|A_{b})\\
            &=\sum_{i=1}^{N_p}w_{b}^i\log(\mu_\theta({s}^i_{bL+1}) l_\theta(o_{bL+1}| s^i_{bL+1})\\
            &\prod_{m=bL+2}^{(b+1)L}g_\theta({s}_m^i| s_{m-1}^i,a_m) l_\theta(o_{m}|{s}^i_{m}))\,\,,
        \end{split}
    \end{equation}
    where $w_b^i$ is the particle weight of the $i$-th particle at block $b$. $\mu_\theta(s^i_{bL+1})$ is the stationary distribution of the particle state $s^i$ at time step $bL+1$. 
    
    Based on the logarithmic pseudo-likelihood function above, we now propose the learning objective for semi-supervised learning as follows:
    \begin{align}
        \theta &=\underset{\theta\in\Theta}{\arg\min}\,\lambda_1 L(\theta)-\lambda_2 Q(\theta)\,\,,\\ \label{eq:semi-loss}
        L(\theta)&=\frac{1}{|T_1|}\sum_{t\in T_1} \delta(s_t^*,\hat{s}_t)\,\,,\\
        Q(\theta)&= \frac{1}{m}\sum_{b=0}^{m-1} \hat{Q}(\theta,\theta_{b})\,\,,
    \end{align}
    where $\lambda_1, \lambda_2$ are the scaling factors depending on the magnitude of components $L(\theta)$ and $Q(\theta)$. $T_1$ is the set of time with true labelled states and $|T_1|$ is the number of elements in $T_1$.  $\delta(s_t^*,\hat{s}_t)$ is a loss function between true labelled states $s_t^*$ and estimated states $\hat{s}_t = \sum_{i=1}^{N_p} w_t^i s_t^i$. $L(\theta)$ is the objective for supervised learning, e.g. the mean squared error (MSE) \cite{karkus2018particle}:
    \begin{equation}
        L(\theta)=\frac{1}{|T_1|}\sum_{t\in T_1} \delta(s_t^*,\hat{s}_t)=\frac{1}{|T_1|}\sum_{t\in T_1}(s_t^*-\hat{s}_t)^T(s_t^*-\hat{s}_t)\,\,,
    \end{equation}
    With the derived semi-supervised learning objective, the pseudocode for the semi-supervised differentiable particle filter (SDPF) framework is presented in Algorithm \ref{alg:alg1}. 
    
    \begin{algorithm}[htpb]
    \caption{Semi-supervised differentiable particle filters framework}
    \label{alg:alg1}
    \SetEndCharOfAlgoLine{}
    \SetKwComment{Comment}{// }{}
    \SetKwInOut{Input}{Input}
    \SetKwFor{For}{for}{do}{end~for}
    \Input{\\\vspace{-1em}\hspace{-2em}\small
    \begin{tabular}[t]{l @{\hspace{.5em}} l l @{\hspace{.5em}} l}%
    $o_{1:T}$ & Observations & $\mu_\theta(s_1)$ & Initial distribution of $s_1$\\
    $a_{1:T}$ & Actions & $T$ & Episode time\\
    $N_p$ & Particle number & $T_1$ & Time step with true label\\
    $\alpha$ & Learning rate & $s_t^*$ & True labelled state\\
    $\lambda_1, \lambda_2$ & Scaling factors & $N_{thres}$ & Resampling threshold\\
    $g_\theta$ & Dynamic model & $L$ & Block length\\
    $l_\theta$ & Meas. model & $f_\theta$ & Action transformer \\
    $d_\theta$ & Cosine distance & $\delta(\cdot)$ & Supervised loss function\\
    $h_\theta$ & Obs. encoder & $\epsilon$ & Reparam. noise vector\\
    $\hat{h}_\theta$ & State encoder & $p(\epsilon)$ & Distribution of $\epsilon$\\
    \end{tabular}\hspace{-0.5em}%
    }
    \BlankLine
    Initialise parameters $\theta$ of $f_\theta$, $h_\theta$ and $\hat{h}_\theta$ randomly; \;
    Draw particles $\{s_1^i\}_{i=1}^{N_p}$ from $\mu_\theta(s_1)$; \;
    Set particle weights  $\{w_1^i\}_{i=1}^{N_p}=\frac{1}{N_p}$; \;
    Set $Q=0$ and $b=0$;\;
    \While{$\theta$ not converged}{
    \For{$k=2$ to $T$ }{
        Compute the effective sample size: $N_{eff}=\frac{1}{\sum_{i=1}^{N_p}(w_{k-1}^i)^2}$;\;
        \uIf{$N_{eff}<N_{thres}$}{
            Select ancestor index $A_{k-1}^i$ with $Pr(A_{k-1}^i=i)=w_{k-1}^i$ for $i=1,...,N_p$;\;
            $\{w_{k-1}^i\}_{i=1}^{N_p}=\frac{1}{N_p}$;\;
        }
        \Else{
            $A_{k-1}^i=i$ for $i=1,...,N_p$;
        }
        $e_k=h_\theta(o_k)$; \;
        \For{$i=1$,...,$N_p$}{
        Draw particles $s_k^i \sim g_\theta(s_k|s_{k-1}^i,a_k)=s_{k-1}^i+f_\theta(s_{k-1}^i, a_k)+\epsilon^i$ where $\epsilon^i\sim p(\epsilon)$; \;
        $\tilde{e}^i_k=\hat{h}_\theta(s_k^i)$, $l^i_k=d_\theta(\tilde{e}^i_k, e_k)$;\;
        $w_k^i=l^i_kw_{k-1}^i$; \;
        Set $s_{1:k}^i=(s_{1:k-1}^{A_{k-1}^i}, s_k^i)$;\;
      }
      Normalize weights $\{w_k^i\}_{i=1}^{N_p}$ so that $\sum_{i=1}^{N_p}w_{k}^i=1$;\;
      \uIf{$k \mod L=0$}{
        \For{$i=1,...,N_p$}{
            $\eta^i=\mu_\theta(s^i_{bL+1})l_\theta(o_{bL+1}|s^i_{bL+1})$\;
            $\,\,\,\,\,\,\prod_{m=bL+2}^{(b+1)L}g_\theta({s}_m^i| s_{m-1}^i,a_m) l_\theta(o_{m}|{s}^i_{m})$;
        }
        Compute $Q=Q+\sum_{i=1}^{N_p}w_{k}^i\log(\eta^i)$;\;
        $b=b+1$;\;
      }
    }
    Total loss: $\mathcal{L}=\lambda_1 \frac{1}{|T_1|}\sum_{t\in T_1} \delta(s_t^*, \sum_{i=1}^{N_p} w_t^i s_t^i)-\lambda_2 \frac{1}{b}Q$ ; \;
    Update model parameters $\theta\leftarrow\theta-\alpha\nabla_\theta \mathcal{L}$; \;
    }
    \end{algorithm}

\section{Experiments}
\label{sec:experiment}

    In this section, we investigate the performance of semi-supervised differentiable particle filters (SDPFs) where ground truth information
    is unavailable for a large portion of time steps\footnote{Code is available at https://github.com/HaoWen-Surrey/SemiDPF.}. The same experiment environments as in \cite{jonschkowski2018differentiable} and \cite{karkus2018particle} with simulated and real-world datasets, namely simulated Maze environment \cite{beattie2016deepmind} and House3D environment \cite{wu2018building}, are used to evaluate algorithms. Since the global map in the House3D environment is unavailable in the Maze environment, to make a fair comparison focusing on the semi-supervised setting, the proposed method is compared with \cite{jonschkowski2018differentiable} and \cite{karkus2018particle} in the Maze environment and the House3D environment, respectively. In both environments, the performance of long short-term memory (LSTM) network \cite{hochreiter1997long} is also compared. 
    
\subsection{Simulated maze environment}
\begin{figure*}[!ht]
    \begin{center}
    \includegraphics[width=1.0\linewidth]{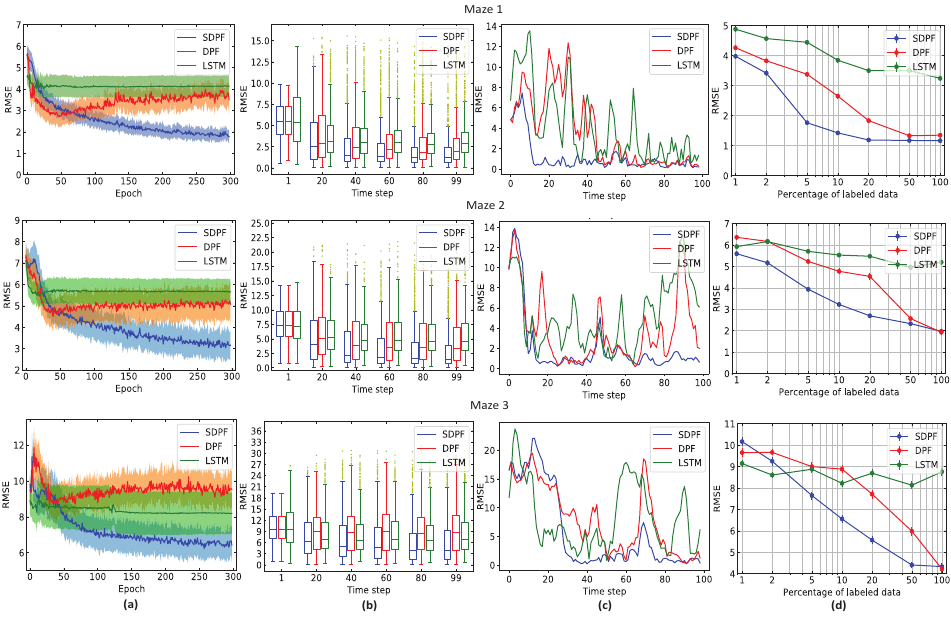}
    \caption{The Root Mean Square Error (RMSE) comparison using the SDPF, the DPF and the LSTM in Maze 1, Maze 2 and Maze 3 (top, middle, bottom row, respectively). (a): The mean and standard deviation of RMSE at the last time step on validation data during training starting from epoch 1. (b): The boxplot of the RMSE at the specified time steps evaluated at 1000 trajectories on testing data. (c): The RMSE of a specified trajectory along the whole time step on testing data. The percentage of labelled data corresponding to results presented in (a), (b) and (c) is $10\%$.  (d): The RMSE for 1000 trajectories on testing data with different percentage of available ground truth data. The error bar represents the standard error of the reported RMSE.}    
    \label{fig:experiment1}
    \end{center}
    \end{figure*}

\subsubsection{Experiment setup}
    In the first experiment, the SDPF is tested on the task of robot \emph{global localisation} in a simulated maze environment. It is modified from the DeepMind Lab environment \cite{beattie2016deepmind} where unique wall textures are removed to make the hidden state to be partially observable. There are three different sizes of mazes in this environment, namely Maze 1, Maze 2, and Maze 3 with the size ranging from small to large. The goal of this task is to estimate the state of the robot from a sequence of visual images and odometry measurements. The robot randomly walks through the maze, and the observation images come from first-person viewpoint of the robot with $32\times32$ RGB image size and the actions data come from the odometry measurements. 
    
    Both the training and testing data consist of 1000 trajectories of 100 steps with one step per second. In the semi-supervised setting, we take 100 trajectories for training with $10\%$ of randomly chosen time steps provided with ground truth state values, and validate on the remaining 900 trajectories. The states $s=[s_x, s_y, s_\beta]$ and actions $a=[a_x, a_y, a_\beta]$ represent the coordinate of position $(x,y)$ and orientation angle $\beta$, and the velocity of translation and rotation, respectively.
    
    The particles are initialised uniformly across the maze. The number of particle used in the SDPF and the DPF is set to be 100. In the dynamic model Eq. \eqref{eq:dynamics_semi}, the actions $a_t$ come from odometry measurements which describe the relative motion in robot coordinate frame, and are transformed into global coordinate frame $f_\theta(s_{t-1}^i, a_t)$. Here, the predicted particles are sampled from the differentiable function $\tilde{s}_t^i=s_{t-1}^i+f_\theta(s_{t-1}^i, a_t)+diag(\sigma_x,\sigma_y,\sigma_\beta)\zeta^i$, where $\zeta^i\sim\mathcal{N}(0;\textit{I})$ is a noise vector from a standard multivariate Gaussian and $I$ is an identity matrix. The standard deviations $\sigma_x$, $\sigma_y$, and $\sigma_\beta$ for translation in $x$ and $y$ axes and the rotation are set to be $20.0$, $20.0$, and $0.5$, respectively. For the SDPF, we choose the block length $L=20$ time steps. Scaling factors $\lambda_1=10$ and $\lambda_2=0.01$. Adam \cite{kingma2014adam} is used as the optimiser and the learning rate is set to $0.0003$.
    
    \subsubsection{Experimental results}
    In this experiment, the error metric is the Root Mean Square Error (RMSE) between the predicted state and the ground truth state, and we scale the RMSE through dividing it by the average step size for a sensible metric across dimensions following the practice in \cite{jonschkowski2018differentiable}. Figure \ref{fig:experiment1} shows the experiment results in the Maze 1, Maze 2 and Maze 3.
    
    \textbf{The SDPF converges to the lowest RMSE during training process.} Figure \ref{fig:experiment1} (a) shows the RMSE at the last time step evaluated on validation data. The SDPF exhibits the smallest RMSE while converges the slowest on validation data among the SDPF, the DPF and the LSTM for all mazes environments. The convergence speed of RMSE for the LSTM is faster but with larger RMSE than that of the DPF on Maze 1 and Maze 2. Among different sizes of maze environments, the SDPF converges to the lowest RMSE.
    
    \textbf{The SDPF significantly improves tracking performance on testing trajectories.} Figure \ref{fig:experiment1} (b) shows the boxplot of RMSE at the specified time step evaluated on testing data. The SDPF has the lowest median RMSE value comparing to the DPF and the LSTM for all mazes environments. Among different sizes of maze environments, the SDPF possesses the lowest median RMSE. Figure \ref{fig:experiment1} (c) shows one representative example of tracking performance, which provides the visualisation comparison of specified trajectory along the whole time steps on testing data in Maze 1, Maze 2 and Maze 3. It shows that the SDPF has the lowest RMSE at a wide range of time steps than the other two algorithms.
    
    \textbf{The SDPF is robust to a wide range of percentage of labelled data.} Figure \ref{fig:experiment1} (d) presents tracking performance of the SDPF, the DPF and the LSTM with different percentage of
    available ground truth state values during training. It shows that the SDPF has the lowest RMSE value under a wide range of percentage of labelled data.

    \begin{figure}[!ht]
    \begin{center}
    \includegraphics[width=0.94\linewidth]{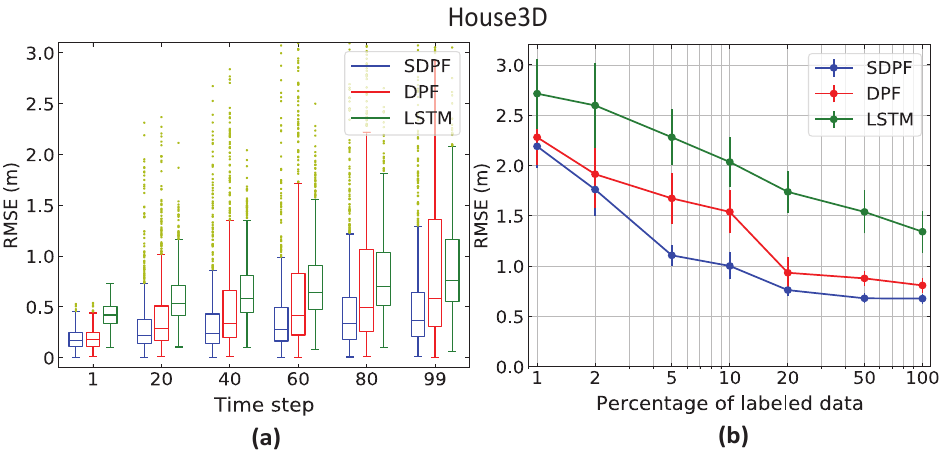}
    \caption{(a) The boxplot of the RMSE at the specified time steps with 10\% of labelled data evaluated at 820 trajectories on the testing data in House3D. (b): The influence of percentage of labelled data on the RMSE of the SDPF, the DPF and the LSTM for all trajectories on testing data. The error bar represents the standard error of the RMSE at specified percentage of labelled data on testing data.}
    \label{fig:experiment2}
    \end{center}
    \end{figure}
    
    \subsection{House3D environment}
    
    \subsubsection{Experiment setup}
    
    The House3D environment \cite{wu2018building} is built upon human-designed residential buildings models from the SUNCG dataset \cite{song2017semantic}. The trajectory comes from the random walk of a robot in the buildings, and the average size of the buildings and rooms are $206\,m^2$ and $37\,m^2$ respectively. The goal of this task is to track the state of the robot under a sequence of global map, visual images, and odometry. The visual images come from a monocular RGB camera with image size of $56\times56$.
    
    The training set is comprised of 74,800 trajectories with 25 time steps, and the validation set contains 830 trajectories with 100 time steps, and the testing set consists of 820 trajectories with 100 time steps from 47 previously unseen buildings. To formulate the semi-supervised setting, we take 1000 trajectories and randomly set the ground truth data to be unknown for training with different labelled ratio. For training and testing of the DPF and the SDPF, the number of particles is set to be 30 and 1000, respectively. The initial particles are sampled from a Gaussian distribution centered at the true initial states with standard deviations $0.30\, m, 0.30\,m, 30^{\circ}$ for translation in $x$ and $y$ axes and the rotation, correspondingly. In dynamic model $\tilde{s}_t^i=s_{t-1}^i+f_\theta(s_{t-1}^i, a_t)+diag(\sigma_x,\sigma_y,\sigma_\beta)\zeta^i$, where $\zeta^i\sim\mathcal{N}(0;\textit{I})$, $\sigma_x,\sigma_y,\sigma_\beta$ are set to be $0.04\, m, 0.04\,m,$ and $5^{\circ}$, respectively. For the SDPF, we choose block length $L=4$ time steps. Scaling factors $\lambda_1=1.0$ and $\lambda_2=0.04$. The optimisation of parameter $\theta$ is based on RMSProp optimiser \cite{tieleman2012lecture} with $0.5$ decay rate, and the initial learning rate is set to be $0.0001$ as in \cite{karkus2018particle}.
    
    \begin{figure}[!ht]
    \begin{center}
    \includegraphics[width=0.9\linewidth]{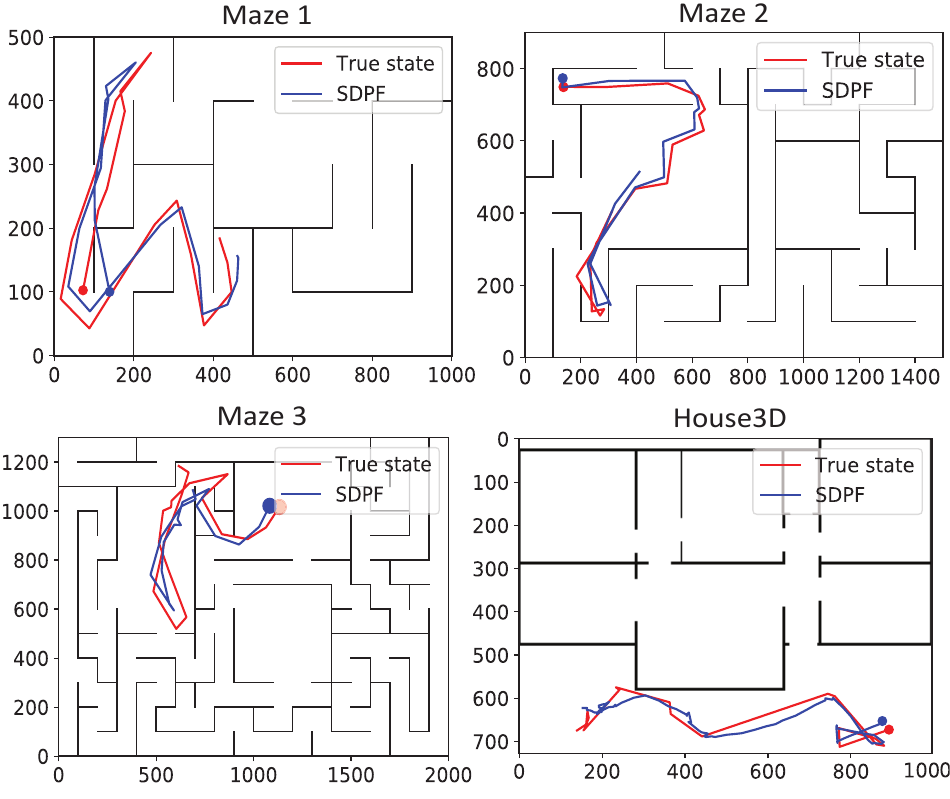}
    \caption{The visualisation of tracking performance of SDPFs in Maze 1, Maze 2, Maze 3 and House3D. For Maze environment, a specified trajectory at the last 20 time steps is presented over Maze 1, Maze 2, Maze 3. For House3D, the trajectory length is 100 time steps. The circle denotes the state at the last time step.}
    \label{fig:VisualTrack}
    \end{center}
    \end{figure}
    
    \subsubsection{Experimental results} We evaluate the performance of SDPFs following the same procedure as in \cite{karkus2018particle}. The main differences between House3D and Maze environments are: (i) The testing environments for House3D are previously unseen environments while they are the same as training environments for Maze. (ii) For the House3D, the likelihood of the predicted particles in measurement model combines features of floor map and observation image, while in the Maze, it is defined by a function of the similarity between features of the observation image and the particle state. Figure \ref{fig:experiment2} shows the comparison between the SDPF, the DPF and the LSTM on the testing set. 
    
    \textbf{The SDPF can generalise to different environments.}  Figure \ref{fig:experiment2} (a) shows the boxplot of RMSE at specified time steps with 10\% of labelled data on the testing data. The SDPF leads to the lowest median RMSE and smallest interquartile range (IQR) value. The SDPF can produce the most accurate tracking results compared with the DPF and the LSTM. Figure \ref{fig:experiment2} (b) shows the RMSE from all trajectories in testing set with different percentage of ground truth samples during training. The SDPF is able to produce significantly lower RMSE compared to the DPF especially when trained with small percentage of labelled data such as 5\% and 10\%. The standard error of RMSE for the SDPF is consistently smaller comparing to the DPF and the LSTM. Figure \ref{fig:VisualTrack} shows a visualisation of the tracking performance of the SDPF in Maze 1, Maze 2, Maze 3 and House3D.
    
\section{Conclusion}
\label{sec:conclusion}

	We propose an end-to-end semi-supervised learning method using maximum pseudo-likelihood estimation for differentiable particle filters (SDPFs), which can leverage unlabelled states and the history of observations and actions to learn the dynamic and measurement models. The proposed SDPF outperforms the DPF and the LSTM in state estimation tasks for both global localisation and tracking when a large portion of ground truth data are unknown.
	
	The semi-supervised differentiable particle filter can be further extended in several directions. For example, invertible neural network can be introduced to replace the conventional neural network and enable flexible transformation while keeping both density estimation and sampling computationally tractable. Another research area is unsupervised learning and self-supervised learning for effective parameter inference in differentiable particle filters without any ground truth data.

\newpage
\balance

\bibliography{example} 
\bibliographystyle{IEEEtran}

\end{document}